\documentclass{article}

\usepackage{arxiv}
\usepackage{graphicx}
\usepackage[utf8]{inputenc} 
\usepackage[T1]{fontenc}    
\usepackage{hyperref}       
\usepackage{url}            
\usepackage{booktabs}       
\usepackage{amsfonts}       
\usepackage{nicefrac}       
\usepackage{microtype}      
\usepackage{lipsum}
\usepackage{amsmath}

\title{Learning Theory and Support Vector Machines \\ a primer}

\author{
  Michael~Banf\\
  EducatedGuess.ai\\
  Siegen, Germany \\
  \texttt{michael@educatedguess.ai} \\
}

\begin{document}
\maketitle

\begin{abstract}
The main goal of statistical learning theory is to provide a fundamental framework for the problem of decision making and model construction based on sets of data. Here, we present a brief introduction to the fundamentals of statistical learning theory, in particular the difference between empirical and structural risk minimization, including one of its most prominent implementations, i.e. the Support Vector Machine. 
\end{abstract}

\keywords{Machine Learning \and Statistical Learning Theory \and Supervised Learning}
\vskip 0.3in

\section*{Introduction to Statistical Learning Theory}

The main goal of statistical learning theory \cite{Vapnik99},\cite{Schoelkopf02} is to provide a fundamental framework for the problem of decision making and model construction based on sets of data. Assumptions can be made of the statistical nature about the underlying phenomena. One of the original problems in applying statistical learning theory is that of binary pattern recognition. Here, given two classes of entities, one has to assign a novel unclassified object to either of the two classes. This problem can be formalized as follow: Suppose we are given $m$ observations, where each observation $i$ consists of the data $\boldsymbol{x}_i \in \mathcal{R}^{n}, i = 1, . . . , m$, as well as a ``ground truth'' labeling $y_i \in \{-1,1\}$. Then, given the data

\begin{equation}
(\boldsymbol{x}_1, y_1),...,(\boldsymbol{x}_m, y_m) \in \mathcal{X} \times \{-1,1\}
\end{equation}

we want to estimate a \textbf{decision function} $f \rightarrow \mathcal{X} \times \{-1,1\}$ that is able to generalize, i.e. avoid model overfitting, to unseen data points, i.e. minimizing the \textbf{expected risk}:

\begin{equation}
R(\alpha) = \int \frac{1}{2} \mid y - f(\boldsymbol{x}, \alpha) \mid d P(\boldsymbol{x},y)
\end{equation}

It is assumed that some unknown probability distribution
$P(y,\boldsymbol{x})$, from which the data is, independently drawn and identically distributed (iid), drawn, does exists. Therefore, if $p(y, \boldsymbol{x})$ exists, $dP(y, \boldsymbol{x})$ may be written as $p(\boldsymbol{x}, y)d\boldsymbol{x}dy$, in order to state the true error rate. Statistical learning theory proves the necessity to restrict the set of functions from which $f$ can be selected with respect to the capacity suitable for the number of available training data, and it provides some upper bound on the test error, depending on the capacity of the function class as well as the empirical risk. Subsequently, minimizing this bound provides the basis for \textbf{structural risk minimization} \cite{Schoelkopf02}. The \textbf{empirical risk} $R_{emp}(\alpha)$ is defined as the measured mean error rate on the training data:

\begin{equation}
R_{emp}(\alpha) = \frac{1}{2m} \sum_{i = 1}^{m} \mid y_i - f(\boldsymbol{x}_i, \alpha) \mid
\label{eq:er}
\end{equation}

Here, $R_{emp}(\alpha)$ is a fixed number for a particular choice of $\alpha$ and a particular training set $\{\boldsymbol{x}_i, y_i \}$. $\frac{1}{2} \mid y_i - f(\boldsymbol{x}_i, \alpha) \mid$ is known as the \textbf{loss function}. 

For binary classification, output values are either 0 or 1. Thus, choosing $\eta$ such that $0 \leq \eta \leq 1$ yields the following bound for the loss function, with probability $1 \: \text{-} \: \eta$ \cite{Schoelkopf02}:

\begin{equation}
R(\alpha) \leq R_{emp}(\alpha) + \sqrt{(\frac{h(\log(2m/h) + 1) - \log(\eta/4)}{m})}
\label{eq:rr}
\end{equation}

$h$ denotes a non-negative integer called the \textbf{Vapnik Chervonenkis (VC) dimension} and describes a property of a set of functions $\{f(\alpha)\}$. It can be defined for various classes of functions $f$. A given set of $m$ observations can be labeled in all possible $2m$ ways, and for each labeling, a member of the set $\{f(\alpha)\}$ can be found that correctly assigns those labels. This is described as ``shattering'' this set of observations by that set of functions. Hence, the VC dimension for such a set of functions $\{f(\alpha)\}$ is defined as the maximum number of training points that can be ``shattered'' by $\{f(\alpha)\}$. Note that if the VC dimension would be $h$, there is at least one particular set of $h$ points that can be shattered. In general, however, it is not the case that each set of $h$ points can be shattered.\\

The second term on the right hand side of equation (\ref{eq:rr}) is called the ``VC confidence'' and the whole right hand side of equation (\ref{eq:rr}) the ``risk bound'', which is independent of $P(\boldsymbol{x}, y)$. The risk bound only presumes that both, the training and test data, are drawn independently according to  $P(\boldsymbol{x}, y)$. Generally, it is not possible to compute the left hand side, however, if $h$ is known, one can compute the right hand side.

\section*{Structural Risk Minimization} 
\addcontentsline{toc}{subsubsection}{Structural Risk Minimization}

\begin{figure*}[h!]
\begin{center}
\includegraphics[width=0.75\textwidth]{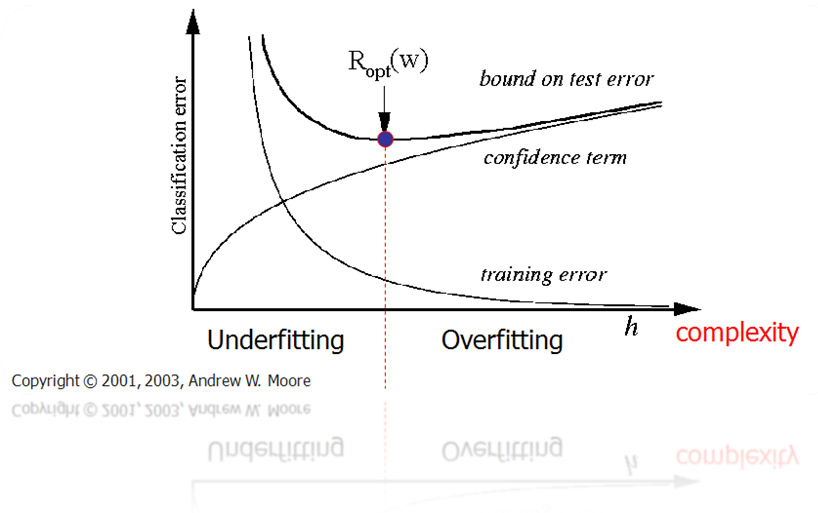}
\caption{Finding the best model complexity to avoid underfitting, i.e. not being able to separate the data appropriately, as well as overfitting, i.e. no being able to generalize to unseen data.}
\label{fig:srm}
\end{center}
\end{figure*}

Thus, given several different families of functions $f(\boldsymbol{x}, \alpha)$, called ``learning machines'' and choosing a fixed, sufficiently small $\eta$, by then taking that machine which minimizes the right hand side, one can choose that machine which provides the lowest upper bound on the actual risk. This provides a method for selecting a learning machine for a given task and is the fundamental idea of structural risk minimization. The VC confidence is described as a monotonic increasing function of $h$ (see in figure \ref{fig:srm}) for any number of observations $m$ and given a selection of learning machines whose empirical risk would be equal to zero, one wants to select a learning machine whose set of functions, associated with it, has a minimal VC dimension, which then leads to a better upper bound on the actual error, as illustrated in figure \ref{fig:srm}. In general, for non zero empirical risk, one selects that specific learning machine that will minimize the right hand side of equation (\ref{eq:rr}). 

\newpage

\section*{Support Vector Machines} 
\addcontentsline{toc}{subsubsection}{Support Vector Machines}

Support Vector Machines can be considered an approximate implementation of the principle of structural risk minimization, as they try to minimize a combination of the empirical risk in equation (\ref{eq:er}), and a capacity term derived for the class of \textbf{hyper-planes} in a dot product space $\mathcal{H}$ \cite{Schoelkopf02},

\begin{equation}
\langle \boldsymbol{w}, \boldsymbol{x} \rangle + b = 0 \textnormal{\quad with \quad} \boldsymbol{w} \in \mathcal{H}, b \in \mathbb{R}
\end{equation}

corresponding to decision functions 

\begin{equation}
f(x) = sgn (\langle \boldsymbol{w}, \boldsymbol{x} \rangle + b).
\label{eq:df} 
\end{equation}

\subsection*{Hard and Soft Margin Solutions}
\addcontentsline{toc}{paragraph}{Hard and Soft Margin Solutions}

\begin{figure*}[t!]
\begin{center}
\includegraphics[width=0.9\textwidth]{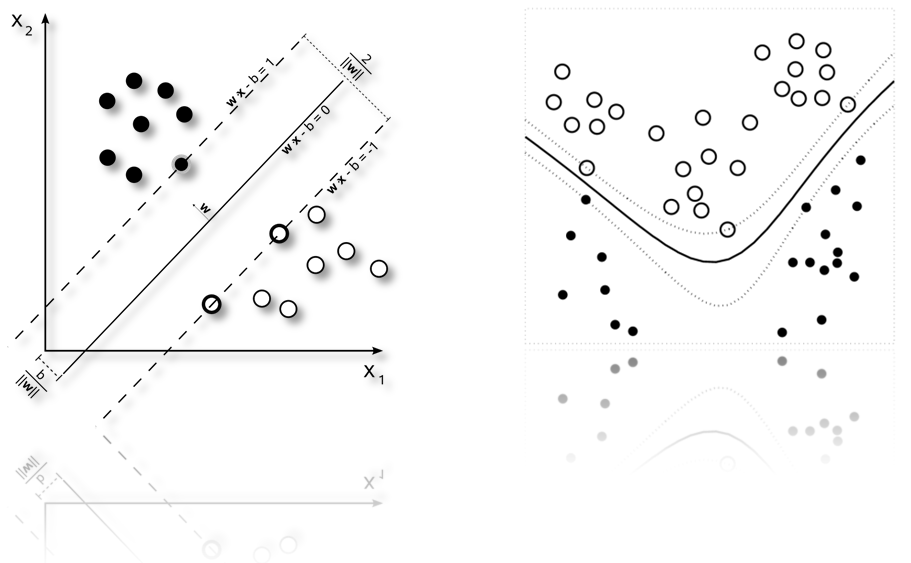}
\caption{Left: Linear separable classification. Optimal hyperplane is shown as a solid
line. Weight vector $\boldsymbol{w}$ and a threshold $b$ yield $y_i ( \langle \boldsymbol{w}, \boldsymbol{x}_i \rangle + b) > 0 \textnormal{\quad} \forall i = 1,...,m$. Support Vectors lie on the borders of the margin (dashed lines). Right: Examples of a non-linear separation surface found using a radial basis function kernel $k(\boldsymbol{x}, \boldsymbol{x'}) = e^{-\Vert \boldsymbol{x} - \boldsymbol{x'} \Vert^2}$.}
\label{fig:lnlsvm}
\end{center}
\end{figure*}

In case of a linearly separable set of observations, a unique optimal hyper-plane exists, differentiated by the maximal margin of separation between any observation point $\boldsymbol{x}_i$ and the hyper-plane, as visualized in figure \ref{fig:lnlsvm} (left). Such a case is called a hard margin solution. This optimal hyper-plane would be the solution of

\begin{equation}
\underset{\boldsymbol{w} \in \mathcal{H}, b \in \mathbb{R}}{maximize} \: \min \lbrace \Vert \boldsymbol{x} - \boldsymbol{x}_i \Vert \mid \boldsymbol{x} \in \mathcal{H}, \langle \boldsymbol{w}, \boldsymbol{x} \rangle + b = 0, i = 1,...,m \rbrace
\end{equation}

Further, the capacity of the class of separating hyper-planes decreases with increasing margin. As illustrated in figure \ref{fig:lnlsvm} (left), in order to construct the optimal hyper-plane, on needs to solve the following quadratic programming problem

\begin{equation}
\underset{\boldsymbol{w} \in \mathcal{H}, b \in \mathbb{R}}{minimize} \: \frac{1}{2} \Vert \boldsymbol{w} \Vert^{2} \newline \textnormal{\quad subject to \quad}  
y_i ( \langle \boldsymbol{w}, \boldsymbol{x}_i \rangle + b) \geq 1 \textnormal{\quad} \forall i = 1,...,m
\label{eq:qpp}
\end{equation}

The constraints ensure that $f(\boldsymbol{x}_i)$ will be $+1$ for $y_i = +1$, and $\text{-}1$ for $y_i = \text{-} 1$. This constrained optimization problem in equation (\ref{eq:qpp}) is computed based on \textbf{Lagrange multipliers} \cite{Burges98} $\alpha_i \geq 0 \: \: (\boldsymbol{\alpha} := (\alpha_1,...,\alpha_m))$ and a \textbf{Lagrangian} 

\begin{equation}
L(\boldsymbol{w},b,\boldsymbol{\alpha}) = \frac{1}{2} \Vert \boldsymbol{w} \Vert^{2} - \sum_{i = 1}^{m} \alpha_i (y_i (\langle \boldsymbol{w}, \boldsymbol{x}_i \rangle + b) - 1)
\label{eq:lagr}
\end{equation}

$L$ has some saddle point in $\boldsymbol{w}$, $b$ and $\boldsymbol{\alpha}$ at the optimal solution of the primal optimization problem. Therefore, it is minimized with respect to the \textbf{primal variables} $\boldsymbol{w}$ and $b$ and maximized with respect to the \textbf{dual variables} $\alpha_i$. Moreover, the product between constraints and Lagrange multipliers in $L$ diminishes at optimality, i.e.,

\begin{equation}
\alpha_i (y_i (\langle \boldsymbol{w}, \boldsymbol{x}_i \rangle + b) - 1) = 0  \textnormal{\quad }  \forall i = 1,...,m
\label{eq:kkt}
\end{equation}

known in optimization theory \cite{Fletcher87} as \textbf{Karush-Kuhn-Tucker conditions} \cite{Fletcher87},\cite{Burges98}. Minimization with respect to the primal variables requires

\begin{equation}
\frac{\partial}{\partial \: b} L(\boldsymbol{w},b,\boldsymbol{\alpha}) = - \sum_{i = 1}^{m} \alpha_i \: y_i = 0 
\label{eq:pv}
\end{equation}

The solution has some expansion in terms of a subset of the observations, i.e. the \textbf{Support Vectors}, with non-zero $\alpha_i$:

\begin{equation}
\frac{\partial}{\partial \: \boldsymbol{w}}  L(\boldsymbol{w},b,\boldsymbol{\alpha}) =  \boldsymbol{w} - \sum_{i = 1}^{m} \alpha_i \: y_i \: \boldsymbol{x}_i = 0 
\label{eq:exp}
\end{equation}

\newpage

Most often, only a fraction of the training examples actually end up being Support Vectors and due to the Karush-Kuhn-Tucker conditions, Support Vectors define the margin (see figure \ref{fig:lnlsvm} (left)). Thus, once all $\alpha_i$ have been found, we can compute $b$ . All remaining training examples $(\boldsymbol{x}_j,y_j)$ turn out to be irrelevant as their constraints $y_j (\langle \boldsymbol{w}, \boldsymbol{x}_j \rangle + b) \geq 1$ can be discarded. Therefore, the hyper-plane is completely determined by the observations closest to them. Substitution of equation (\ref{eq:pv}) and equation (\ref{eq:exp}) by the Lagrangian in equation (\ref{eq:lagr}) eliminates the primal variables $\boldsymbol{w}$ and $b$, resulting in the \textbf{dual optimization problem}:

\begin{equation}
\underset{\alpha \in \mathbb{R}^{m}}{maximize} \sum_{i = 1}^{m} \alpha_i -  \frac{1}{2} 
\sum_{i,j = 1}^{m} \alpha_i \: \alpha_j \: y_i \: y_j \: K_{ij}
\textnormal{\quad subject to \quad} \alpha_i \geq 0 \quad \forall i = 1,...,m \: \wedge \: \sum_{i = 1}^{m} \alpha_i \: y_i = 0
\label{eq:dop}
\end{equation}

with $K_{ij} := \langle \boldsymbol{x}_i, \boldsymbol{x}_j \rangle$. Based on equation (\ref{eq:exp}), the decision function in equation (\ref{eq:df}) is written as

\begin{equation}
f(\boldsymbol{x}) = sgn \left( \sum_{i = 1}^{m} y_i \: \alpha_i \: \langle \boldsymbol{x}, \boldsymbol{x}_i \rangle + b \right)
\end{equation}

Here, $b$ can be computed using equation (\ref{eq:kkt}). In practise, however, a separating hyper-plane may not exist, e.g. if noise within the training data causes a large overlap of the classes. In consideration of such a case, \textbf{slack variables}  $\xi_i \geq 0 \: \forall
i = 1, . . . ,m$ are introduced in order to relax the constraints of equation (\ref{eq:qpp}) to

\begin{equation}
y_i (\langle \boldsymbol{w}, \boldsymbol{x}_i \rangle + b) \geq 1 - \xi_i \quad \forall i = 1,...,m
\label{eq:gpp2}
\end{equation}

\newpage

Thus, a learning machine that generalizes appropriately is found by controlling both, the classifier capacity, based on $\Vert \boldsymbol{w} \Vert$, and the sum of the slack variables $\sum_{i = 1}^{m} \xi_i$, which provides an upper bound on the number of training errors. Such a classifier, known as \textbf{soft margin solutions}, is obtained by minimizing the objective function

\begin{equation}
\frac{1}{2} \Vert \boldsymbol{w} \Vert^{2} +  C \: \sum_{i = 1}^{m} \xi_i
\end{equation}

subject to the constraints on $\xi_i$ and equation (\ref{eq:gpp2}). The constant $C > 0$ determines the trade-off between margin maximization and training error minimization. Again, this leads to the problem of maximizing equation (\ref{eq:dop}), subject to modified constraint with the only difference from the separable case being an upper bound $C$ on the Lagrange multipliers $\alpha_i$. Another realization replaces the parameter $C$ by a parameter $\nu \in (0,1]$ that provides upper and lower bounds for the subset of examples which become Support Vectors and those which will have non-zero slack variables, respectively \cite{Schoelkopf02}. 

\subsection*{Non Linear Support Vector Machines}
\addcontentsline{toc}{paragraph}{Non Linear Support Vector Machines}

If the decision function $f$ in equation (\ref{eq:df}) is not linear, all above methods have to be generalized to these cases. Boser \textit{et al.} \cite{Boser92}, proved that a rather old method \cite{aizerman1964}, referred to as \textbf{kernel trick}, can be used to accomplish this in a straightforward manner. Thereby, symmetric similarity measures of the form $k: \mathcal{H} \times \mathcal{H} \rightarrow \mathcal{R}$, with $(\boldsymbol{x},\boldsymbol{x'}) \rightarrow k(\boldsymbol{x},\boldsymbol{x'})$, are considered. These functions, given two observations $x$ and $x'$ return a real number value denoting their similarity. To allow for a variety of similarity measures and learning algorithms, the observations are represented as vectors in an arbitrary selected \textbf{feature space} $\boldsymbol{\Phi}$, due to the mapping: $\boldsymbol{\Phi} = \mathcal{X} \rightarrow \mathcal{H}$ with $\boldsymbol{x} \rightarrow \boldsymbol{\Phi}(\boldsymbol{x})$. The function $k$ is often referred to as a \textbf{kernel}. Some popular kernel choice are \textbf{Gaussian}, $k(\boldsymbol{x},\boldsymbol{x'}) = e^{- \frac{\Vert \boldsymbol{x} - \boldsymbol{x'} \Vert^2}{2 \sigma^2}}$, or \textbf{Radial Basis Functions}, $k(\boldsymbol{x},\boldsymbol{x'}) = e^{\gamma \Vert \boldsymbol{x} - \boldsymbol{x'} \Vert^2}$, as shown in figure \ref{fig:lnlsvm} (right). Finally, $f$ can be rewritten as

\begin{equation}
f(x) = sgn \left( \sum_{i = 1}^{m} y_i \: \alpha_i \: k(\boldsymbol{x},\boldsymbol{x_i}) + b \right)
\end{equation}

Furthermore, in the quadratic optimization problem (\ref{eq:dop}) the definition of
$K_{ij}$ becomes $K_{ij} = k(\boldsymbol{x_i}, \boldsymbol{x_j})$.

\subsection*{Probability Estimates} 
\addcontentsline{toc}{paragraph}{Probability Estimates}

As discussed, Support Vector Machines predict only class labels without probability information. However, several extensions to provide probability estimates have been presented \cite{Wu04},\cite{Platt99}. Briefly, given $n$ classes of data, for any observation $\boldsymbol{x}$, the goal would be to estimate

\begin{equation}
p_i = P(y = i| \boldsymbol{x}), \: i = 1,...,n
\end{equation}

Following \cite{Wu04},\cite{Platt99}, in the context of the one-against-one (i.e., pairwise) approach for multi-class classification, we first estimate pairwise class probabilities as 

\begin{equation}
r_{ij} \approx P(y = i| y = i\: \wedge \: j, \boldsymbol{x})
\end{equation}

If $\widehat{f}$ is the decision value at $\boldsymbol{x}$, then we assume

\begin{equation}
r_{ij} \approx \frac{1}{1 + e^{A\:\widehat{f} + B}}
\label{eq:pe}
\end{equation}

where $A$ and $B$ are estimated by minimizing the negative log likelihood of training data, using their labels and decision values. It has been observed that decision values from training may overfit the model in equation (\ref{eq:pe}), so cross-validation is conducted to obtain decision values before minimizing the negative log likelihood. After collecting all $r_{ij}$ values, Wu \textit{et al.} \cite{Wu04} propose various possible approaches to obtain $p_i, \: \forall i$.

\subsection*{Hyper-parameter Estimation}
\addcontentsline{toc}{paragraph}{Parameter Estimation}

Also there exist  a varity of sophisticated approaches to infer the best set of hyper-parameters, i.e. the most appropriate parameters $C$ for linear Support Vector Machines, as well as $(C,\gamma)$, for Radial basis functions based non-linear Support Vector Machines, would be grid search based cross validation.

\bibliographystyle{unsrt}  


\end{document}